\pdfoutput=1

\documentclass[11pt]{article}

\usepackage[preprint]{coling}

\usepackage{times}
\usepackage{latexsym}

\usepackage[T1]{fontenc}

\usepackage[utf8]{inputenc}

\usepackage{microtype}

\usepackage{inconsolata}

\usepackage{graphicx}
\usepackage{multirow}
\usepackage{subcaption}
\usepackage{url}
\title{STAYKATE: Hybrid In-Context Example Selection Combining Representativeness Sampling and Retrieval-based Approach - A Case Study on Science Domains}

\author{Chencheng Zhu\\
Kyushu Institute of Technology \\
\texttt{zhu.chencheng822@mail.kyutech.jp}
\And
Kazutaka Shimada \\
Kyushu Institute of Technology \\
\texttt{shimada@ai.kyutech.ac.jp}
\AND
Tomoki Taniguchi \\
Asahi Kasei \\
\texttt{taniguchi.tcr@om.asahi-kasei.co.jp} \And
Tomoko Ohkuma \\
Asahi Kasei \\
\texttt{okuma.td@om.asahi-kasei.co.jp}
}

\begin{document}
\maketitle
\begin{abstract}
Large language models (LLMs) demonstrate the ability to learn in-context, offering a potential solution for scientific information extraction, which often contends with challenges such as insufficient training data and the high cost of annotation processes. 
Given that the selection of in-context examples can significantly impact performance, it is crucial to design a proper method to sample the efficient ones.
In this paper, we propose STAYKATE, a static-dynamic hybrid selection method that combines the principles of representativeness sampling from active learning with the prevalent retrieval-based approach. 
The results across three domain-specific datasets indicate that STAYKATE outperforms both the traditional supervised methods and existing selection methods. 
The enhancement in performance is particularly pronounced for entity types that other methods pose challenges. 
\end{abstract}

\section{Introduction}
The number of published scientific research has increased exponentially over the past several decades. 
However, valuable data are often buried deep in extensive publications, making it challenging for researchers to retrieve relevant information efficiently \cite{Hong2021}. 
In this situation, Named Entity Recognition (NER) plays a crucial role by identifying specific elements such as names of materials, genes, and diseases within vast unstructured text \cite{Weston2019NamedER}. 
Recent advancements in pre-trained language models (PLMs), encompassing both general-domain and scientific-specific ones, have demonstrated considerable abilities in scientific NER \cite{bert, scibert, clinical-bert, matbert, matscibert}. 
In addition, the rise of large language models (LLMs), such as GPT-3 \cite{gpt-3} and Llama 3 \cite{llama}, has shown their impressive performance on various NLP tasks \cite{systematic}. 
Some studies have reported that automatic annotation using LLMs achieves nearly human-level accuracy
\cite{goodannotator, gptoutperform}. 
Pretrained-based approaches generally require a large amount of high-quality labeled data.
Moreover, LLMs can adapt to new tasks and domains through limited demonstrations without updating their parameters. 
This paradigm, termed in-context learning (ICL) and initially introduced by \cite{gpt-3}, has garnered attention due to its potential for substantial reductions in data annotation requirements.
However, ICL is sensitive to the provided in-context examples. 
Substantial performance variance has been observed when different randomly selected demonstrations are provided in the prompt \cite{kate, zhang-active, towards}.
Therefore, a series of in-context example selection strategies have been proposed to bring out the full potential of ICL. 

\begin{figure*}[t]
  \centering
  \includegraphics[width=15cm]{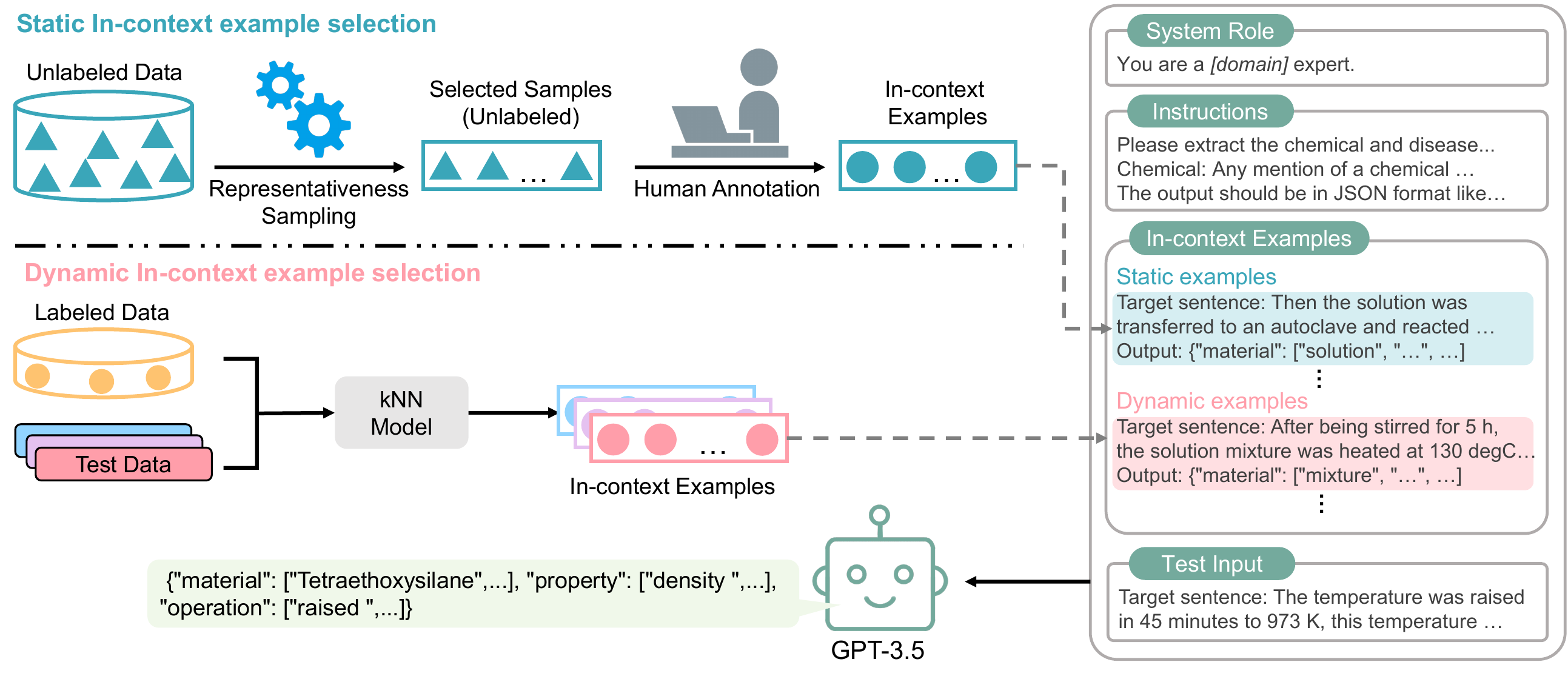}
  \caption{The overall process of STAYKATE. The right side of the figure shows the entire prompt structure.}
  \label{fig:workflow}
\end{figure*}

One approach is to select some examples from unlabeled data and to annotate them by human annotators.
Since these examples remain constant and unchanged across the test set, we refer to this approach as ``static'' in this paper.
In this situation, we need to select efficient examples adequately represent the entire dataset \cite{casestudy}.
In addition, the annotation process requires domain experts for the scientific domain \cite{clinical-manual-prompt} and is costly.
Therefore, we need to select fewer representative examples for the annotation process.
As a result, this approach has the drawback of limiting the number of selected examples for ICL due to the cost.

Another approach is to select some examples similar to the test examples.
This approach is called ``dynamic'' in this paper because the selected examples vary with the test examples.
Some researchers have reported approaches for retrieving semantically similar samples for each test input from the training set \cite{kate, rubin}. 
\citet{kate} proposed a dynamic example selection strategy follows one of the most prevalent retrieval-based methods, KATE (\textbf{K}nn-\textbf{A}ugmented in-con\textbf{T}ext \textbf{E}xample selection). 
Here, we consider the problems specific to scientific domains.
On the one hand, the size of data in scientific domains is not always abundant, namely low-resource setting.
KATE selects the nearest example to the test cases.
In the low-resource setting, it is not always similar to the test example, even if the example is the closest to the test example. 
In other words, there may not be any similar examples due to limited resources for the retrieval process.
On the other hand, the prevalence of multiple synonyms and abbreviations for entities in scientific literature exacerbates the challenge of recognition. 
Moreover, scientific NER is more context-sensitive \cite{biomedicine}. 
As a result, those similar demonstrations may mislead LLMs to derive shallow clues from surface representations \cite{gptre}, while the entities may possess different interpretations under subtle nuances.
Therefore, relying solely on this approach is insufficient.

To resolve the issues from each approach, in this paper, we integrate the two approaches.
The static example selection strategy is inspired by conventional active learning frameworks. 
The objective of active learning is to direct the resource-intensive labeling process toward the most informative and representative instances from a large unlabeled data pool \cite{activelearning}.
This strategy minimizes the cost of annotation while maximizing the utility of unlabeled data. 
The dynamic example selection strategy is KATE. 
We name the hybrid method STAYKATE (\textbf{STA}tic\&d\textbf{Y}namic \textbf{K}nn-\textbf{A}ugmented in-con\textbf{T}ext \textbf{E}xample selection).
Figure \ref{fig:workflow} shows the overall workflow of STAYKATE.
In our experiment, we conduct evaluations across three entity extraction datasets from different scientific subdomains: MSPT \cite{mspt} as materials science, WLP \cite{wlp} as biology, and BC5CDR \cite{bc5cdr} as biomedicine. 
The experimental results show that STAYKATE performs best among the in-context example selection methods examined.
Additionally, our error analysis reveals that STAYKATE not only mitigates common errors such as overpredicting but also enhances the model's ability to discern subtle nuances in the data.

\section{Related work}
\subsection{NER for scientific literature}
In recent years, research on information extraction in the scientific domain has focused on the attributes of textual data. 
One notable issue in scientific literature is the inconsistency in the notation of compounds. 
Handling the partial structures of compounds and the unique variations in their notation as they are presented proves to be challenging. 
In response to this challenge, \citet{watanabe} proposed a multi-task learning approach utilizing a NER model combined with a paraphrasing model. 
Language models tailored specifically for the science domain, such as SciBERT \cite{scibert}, BioBERT \cite{biobert}, and MatSciBERT \cite{matscibert}, have also been proposed. 
These models showed high performance in tasks related to the science domain but require fine-tuning with highly specialized datasets. 
\citet{matsci} indicated that the performance of these models dramatically decreases in low-resource settings.

\subsection{In-context learning}
A novel paradigm known as in-context learning (ICL) \cite{gpt-3} has emerged as a substantial advancement in large language models (LLMs). 
Compared with the traditional fine-tuned base models that need task-specific supervised data and parameter updates, ICL enables LLMs to perform various tasks using a limited set of ``input-output'' pairs as demonstrations, without necessitating any retraining. 
However, prior studies \cite{kate, zhang-active, towards} have shown that the ICL performance is highly sensitive to the given in-context examples. 
To enhance the ICL, a series of strategies have been proposed, including optimizing the order of examples \cite{ordering}, utilizing the most confident model outputs \cite{confidence}, and selecting the most informative data \cite{towards}. 
Another commonly applied solution involved dynamically selecting demonstrations through retrieval modules. 
\citet{kate} have proposed KATE, employed the kNN to retrieve the semantically similar data point in the training set to serve as the in-context examples. 

\subsection{In-context learning for NER}
Several studies have indicated that LLMs may not be optimally suited to NER tasks. 
Compared with tasks like classification and question-answering which align more closely with the pre-training objective of generating natural text, NER presents additional challenges due to its need for intricate span extraction and precise type classification \cite{learningner}. 
This issue is exacerbated in the context of scientific NER. 
\citet{thinking-bio} have indicated that GPT-3 significantly underperforms small PLMs, even when techniques such as contextual calibration and dynamic in-context example retrieval are employed.

\section{Proposed method: STAYKATE}
In the following, we first introduce our prompt construction. 
Then, we elaborate on both the static and dynamic example selection.

\subsection{Prompt Construction}
As illustrated in the right side in Figure \ref{fig:workflow}, the prompt is composed of four components: system role, task instructions, in-context examples, and test input.

\citet{sysrole} demonstrated that specifying system roles for GPT models results in enhanced performance on specific tasks. 
In this research, we adopt a template such as “You are a/an \textit{[domain]} expert.” to define the system role.
The task instructions provide a description of NER task, brief definitions for each target entity, and the required output format. Where available, entity definitions are taken from the dataset’s annotation manual; otherwise, we ask GPT-3.5 to generate definitions based on its own understanding. 
In regard to the output format, GPT-3.5 is instructed to produce responses in JSON format.
In-context examples combine static and dynamic examples. 
The static examples are fixed. They are unchanged across the test set. In contrast, the dynamic examples are different for each test input. They are selected based on their similarity to the test input.
Finally, the test input is provided to GPT-3.5 for generating the response.

\subsection{Static Example Selection}
\subsubsection{Problem Setting}
For static example selection, the goal is to select the most representative examples from a large pool of unlabeled data $D_{unlab}$. 
Given the close resemblance between this problem setting and the active learning (AL) framework, we introduce one of the most prevalent AL strategies, representativeness sampling, to select $k_s$ data points as static samples. 
It is assumed that the selected $k_s$ unlabeled samples will subsequently be annotated by domain experts. 
The cost of human annotation is minimal, given that the number of samples selected is typically limited to a small number (from 2 to 8 in our experiment). 

\subsubsection{Representativeness Sampling}
Predictive Entropy is a metric for quantifying the uncertainty inherent in the model’s predictions. 
By measuring the dispersion of predicted probabilities across all possible label classes for each token, predictive entropy offers insight into the model's confidence in its prediction. 
In our context, predictive entropy is employed for the purpose of assessing the representativeness of the example candidates in unlabeled data pool. 
We first fine-tuned a BERT model under the low-resource setting, denoted as $M$ to obtain the probability $p(y_{i}|x_{i}) = M(x_{i})$ for each token $x_{i}$ of a candidate sentence $x$. 
Then, we define the token level predictive entropy $PE(x_{i})$ as:
\begin{equation}
\label{eq1}
PE(x_{i}) = -\sum_{j=1}^{C} p(y_{i} = c_{j} \mid x_{i}) \log p(y_{i} = c_{j} \mid x_{i})
\end{equation}
where $c_{1}, c_{2}, . . . c_{C}$ are the class labels.

Subsequently, the predictive entropy of each candidate sentence can be defined as:
\begin{equation}
\label{eq2}
H(x) = \frac{1}{N_x} \sum_{i=1}^{N_x} PE(x_i)
\end{equation}
where $N_x$ is the number of tokens in the candidate sentence $x$.
Selecting sentences with exceedingly low $H(x)$ is unreasonable as they can be very simple samples. 
Moreover, sentences with exceedingly high $H(x)$ can be either noise or out-of-domain data. 
Such sentences may prove detrimental to in-context learning. 
Inspired by \cite{kumar-diversity} work, we define a representativeness score, denoted as $R_{Score}$, for selecting the $k_s$ most representative sentences. 
Specifically,
\begin{equation}
\label{eq3}
R_{Score} = |H(x)- (\mu_H + \lambda * \sigma_H)| 
\end{equation}
where $\mu_H$ is mean and $\sigma_H$ is standard deviation of the predictive entropy of all sentences in the unlabeled data pool. 
It is assumed that the selection around $\mu_H$ can represent the overall patterns of the corpus. 
Additionally, $\sigma_H$ is introduced to reduce the possible bias caused by the characteristics of the corpus. 
Therefore, $\lambda$ is introduced to control the distance of the preferred selection zone from $\mu_H$. 
We select $k_s$ candidates which have the lowest $R_{Score}$ as the static in-context examples.



\subsection{Dynamic example selection}
Many studies have indicated that retrieving semantically similar examples for a test input can significantly enhance the in-context learning performance of GPT-3.5. 
In this research, we follow one of the retrieval-based methods, KATE (Knn-Augmented in-conText Example selection), to select the examples dynamically for each test input.

It should be noted that unlike static example selection, which samples from a large pool of unlabeled data $D_{unlab}$, dynamic selection samples from a limited pool of labeled data $D_{lab}$. 
We utilize the OpenAI embedding model to convert both labeled data and test data into embeddings. 
Subsequently, for each test input $s$, the nearest $k_d$ neighbors, $d_1, d_2, …, d_{k_d}$, are retrieved from the $D_{lab} = \{d_i\}_{i=1}^{N_i}$ based on their distances in the embedding space. 
Given a pre-defined similarity measure $sim$ such as cosine similarity, the neighbors are ranked in order such that $sim(d_i, s) \le sim(d_j , s)$ when $i < j$.


\section{Experiment}
\subsection{Datasets} 
We evaluate on three scientific literature corpora for NER tasks from different subdomains.
Due to the running cost of GPT-3.5 usage, we sampled approximately 200 sentences from the original test sets for WLP and BC5CDR. 
The distributions of the datasets are provided in Appendix \ref{app:data_statistics}.

\paragraph{MSPT}
The Material Science Procedural Text corpus \cite{mspt} consists of 230 synthesis procedures annotated by domain experts. 
Although it contains 21 entity types, we focus on the three most important for evaluation: \texttt{Material}, \texttt{Operation}, and \texttt{Property}\footnote{In MSPT, the distinctions between material-descriptor and property-misc are too subtle to discern without the full context of the article. Since GPT-3.5 responses at the sentence level, we merge the \texttt{Material-descriptor} and \texttt{Property-misc} into a \texttt{Property} label.} for evaluation. 

\paragraph{WLP} The Wet Lab Protocol corpus \cite{wlp} contains 622 protocols for biology and chemistry experiments annotated by 10 annotators. 
There are 18 entity types in total. 
We evaluate on \texttt{Actions} and 4 types of Object Based Entities: \texttt{Reagent}, \texttt{Location}, \texttt{Device}, and \texttt{Seal}. 

\paragraph{BC5CDR} The BioCreative V Chemical-Disease Relation corpus \cite{bc5cdr} comprises 1,500 PubMed abstracts. 
\texttt{Chemical (Drug)} and \texttt{Disease} entities are annotated with the annotators who had a medical training background and curation experience. 
We evaluate both types of entities.

\subsection{Baseline Methods}
\paragraph{Fine-tuned BERT}
To compare GPT-3.5 with conventional supervised learning, we trained a NER model under a low-resource scenario using BERT \cite{bert}. 
These fine-tuned BERT models were also utilized in the static example selection strategy to obtain the predicted probabilities for unlabeled example candidates.

\paragraph{Zero-shot}
In the zero-shot setup, we keep the system role, task instructions, and test input in the prompt.

\paragraph{Random Sampling}
We randomly selected $k$ in-context examples from the labeled data pool.

\begin{figure}[t]
  \centering
  \includegraphics[width=6.5cm]{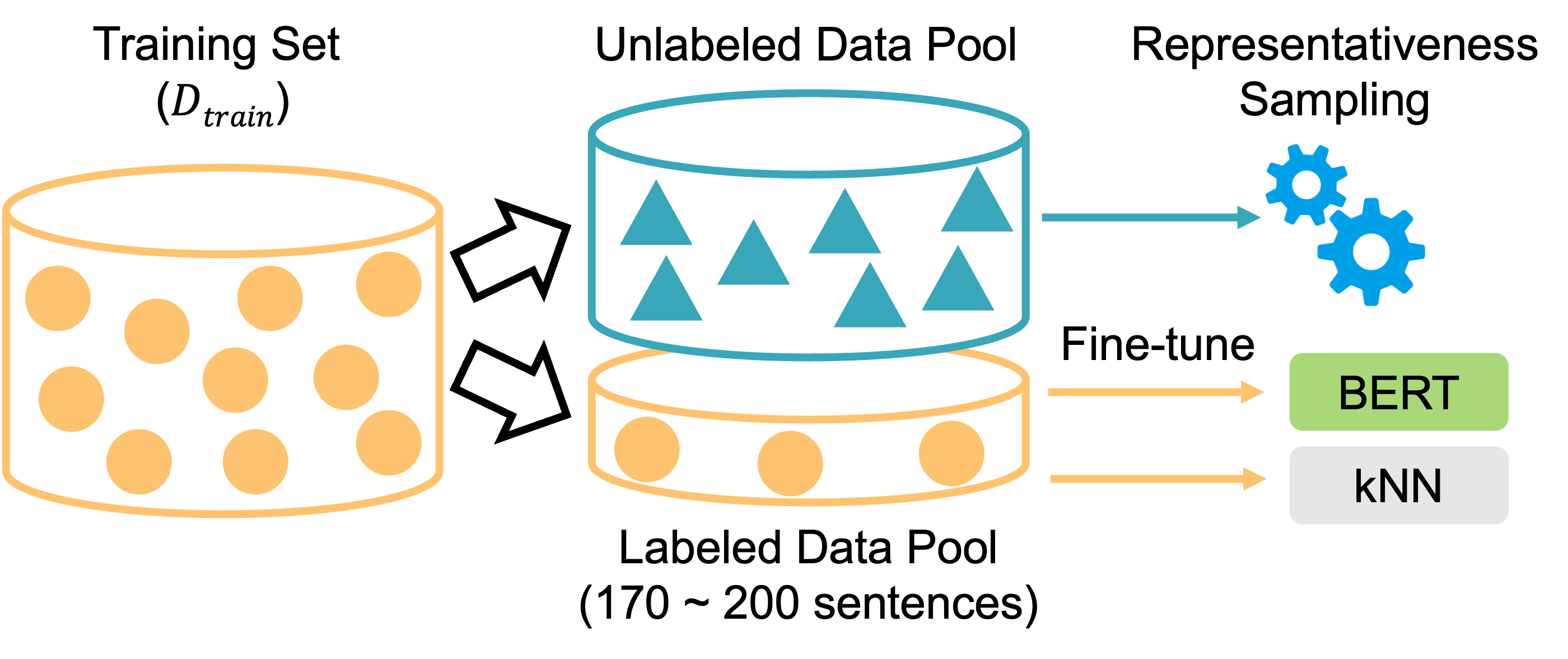}
  \caption{Our experimental setting about data pool.}
  \label{fig:setting}
\end{figure}

\subsection{Experimental Setup}
For each dataset, we divided the original training set ($D_{train}$) into a labeled data pool and a unlabeled data pool.  
We set the labeled data pool consisting of 170 to 200 sentences randomly extracted from the $D_{train}$. 
In a real-world situation, this range can be considered as a reasonable estimate of annotated data for scientific literature.
This labeled data pool is employed for fine-tuning BERT models. 
While the in-context examples selected by KATE are retrieved from the same pool, the static in-context examples are selected from the unlabeled data pool, which is the rest of training set, with labels removed. 
In other words, although the static selection in Figure \ref{fig:workflow} uses unlabeled data, we use the rest of training data that were not used by the dynamic selection in the experiment (See Figure \ref{fig:setting}).
We run all experiments using three distinct data pools and report the mean to mitigate the impact of data variability.

All BERT models were fine-tuned using the HuggingFace\footnote{\url{https://huggingface.co/google-bert/bert-base-uncased}} Transformers library. 
The hyperparameters we used are given in Appendix \ref{app:hyper}.
On account of low-resource setting, we only used 170 to 200 annotated sentences for training. 
For validation data, we used GPT-3.5 to generate the pseudo labels for approximately 200 sentences randomly extracted from the original validation sets. 
The prompts provided to the GPT-3.5 include six random in-context examples.

The GPT model employed in this research was gpt3.5-turbo-16k-0613\footnote{\url{https://platform.openai.com/docs/models/gpt-3-5-turbo}}. The temperature is set to 0.
For dynamic example selection, we used OpenAI embedding model, text-embedding-3-small\footnote{\url{https://platform.openai.com/docs/guides/embeddings}}, to embed labeled data and test input. The embedding dimension is 1536.

In our evaluation, we utilize $k = 2, 6, 8$ in-context examples. In the case of STAYKATE, we allocate $k_s$ for static and $k_d$ for dynamic respectively, where $k = k_s + k_d$. Specifically, for $k = 2$, we set $k_s = k_d = 1$; for $k = 6$ and $8$, $k_s = 2$ and $k_d = 4$ or $6$, respectively.

When calculating the $R_{Score}$ used in the static example selection method, we try either the parameter $\lambda = 0$, or $\lambda = 1$.
We utilize entity-level F1 scores for individual entity types and micro F1 scores for overall evaluation.

\section{Results \& Discuss}
\subsection{Main Results}
The main experimental results are given in Table \ref{tab:overall_result}, Table \ref{tab:entitylevel_result}, and Table \ref{tab:nshot_result}. 
Besides the baseline methods and our proposed method, we also report the results using only static examples (Representativeness) and only dynamic examples (KATE).


\begin{table}[t]
\scalebox{0.86}{
\begin{tabular}{lccc}
\hline
\textbf{Method}             & \textbf{MSPT} & \textbf{WLP}  & \textbf{BC5CDR} \\ \hline
\multicolumn{4}{l}{\textit{Baseline Methods}}                                 \\
BERT (Low-resource) & 0.55          & 0.37          & 0.53            \\
Zero-shot                   & 0.42          & 0.50          & 0.62            \\
Random Sampling             & 0.57          & 0.65          & 0.69            \\ \hline
\multicolumn{4}{l}{\textit{In-context example selecting methods}}              \\
Representative ($\lambda = 0$)  & 0.57          & 0.66          & 0.69            \\
Representative ($\lambda = 1$)  & 0.54          & 0.65          & 0.69            \\
KATE                        & 0.60          & 0.68          & 0.70            \\ \hline\hline
\multicolumn{4}{l}{\textit{\textbf{Ours}}}                                    \\
STAYKATE ($\lambda = 0$)            & \textbf{0.61} & \textbf{0.69} & 0.70            \\
STAYKATE ($\lambda = 1$)            & 0.60          & 0.66          & \textbf{0.72}   \\ \hline
\end{tabular}
}
\caption{Overall results for baseline methods and in-context example selecting methods. The best results are given in \textbf{bold}. We run all experiments three times with different data pool and report the mean of micro average F1.}
\label{tab:overall_result}
\end{table}

\begin{table*}[t]
\centering
\small
\begin{tabular}{cl|ccc|cccccc}
\hline
\multicolumn{1}{l}{}             &                           & \multicolumn{3}{c|}{\textbf{Baseline Methods}} & \multicolumn{6}{c}{\textbf{In-context Example Selecting Methods}}                           \\ \hline
\multicolumn{1}{l}{}             &                           & BERT        & Zero-shot        & Random        & \multicolumn{2}{c}{Representative} & KATE          & Random & \multicolumn{2}{c}{STAYKATE}  \\
                                 &                           &             &                  &               & $\lambda = 0$        & $\lambda = 1$              &               & + KATE & $\lambda = 0$         & $\lambda = 1$        \\ \hline
\textbf{Dataset}                 & \textbf{Entity (support)} &             &                  &               &              &                     &               &        &               &               \\
\multirow{4}{*}{\textbf{MSPT}}   & Material (338)            & 0.51        & 0.49             & 0.56          & 0.58         & 0.54                & 0.60          & 0.59   & \textbf{0.62} & 0.60          \\
                                 & Operation (242)           & 0.75        & 0.51             & 0.76          & 0.74         & 0.75                & \textbf{0.77} & \textbf{0.77}   & 0.76          & \textbf{0.77} \\
                                 & Property (105)            & 0.14        & 0.05             & 0.15          & 0.17         & 0.17                & 0.22          & 0.21   & \textbf{0.25} & 0.20          \\  
                                 & micro avg                 & 0.55        & 0.42             & 0.57          & 0.57         & 0.54                & 0.60          & 0.60   & \textbf{0.61} & 0.60          \\ \hline
\multirow{6}{*}{\textbf{WLP}}    & Action (275)              & 0.57        & 0.76             & 0.79          & 0.80         & 0.81                & \textbf{0.82} & 0.80   & 0.81          & 0.80          \\
                                 & Device (45)               & 0.00        & 0.10             & 0.24          & 0.21         & 0.25                & 0.25          & 0.27   & \textbf{0.29} & 0.22          \\
                                 & Location (122)            & 0.00        & 0.39             & 0.54          & 0.57         & 0.51                & \textbf{0.57} & 0.54   & 0.55          & 0.52          \\
                                 & Reagent (178)             & 0.22        & 0.39             & 0.62          & 0.64         & 0.61                & 0.66          & 0.65   & \textbf{0.67} & 0.63          \\
                                 & Seal (20)                 & 0.00        & 0.26             & 0.54          & 0.43         & \textbf{0.56}       & 0.40          & 0.40   & 0.48          & 0.41          \\ 
                                 & micro avg                 & 0.37        & 0.50             & 0.65          & 0.66         & 0.65                & 0.68          & 0.67   & \textbf{0.69} & 0.66          \\ \hline
\multirow{3}{*}{\textbf{BC5CDR}} & Chemical (291)            & 0.64        & 0.63             & 0.71          & 0.70         & 0.71                & 0.72          & 0.71   & 0.72          & \textbf{0.74} \\
                                 & Disease (185)             & 0.36        & 0.62             & 0.65          & 0.66         & 0.67                & 0.67          & 0.67   & 0.67          & \textbf{0.70} \\
                                 & micro avg                 & 0.53        & 0.62             & 0.69          & 0.69         & 0.69                & 0.70          & 0.69   & 0.70          & \textbf{0.72} \\ \hline
\end{tabular}
\caption{Experimental results on entity-level for each dataset. The best results are given in \textbf{bold}. The value of micro average for each dataset can be found the same in Table \ref{tab:overall_result}.}
\label{tab:entitylevel_result}
\end{table*}

\begin{figure*}[t]
  \begin{subfigure}{0.31\textwidth}
    \includegraphics[width=\linewidth]{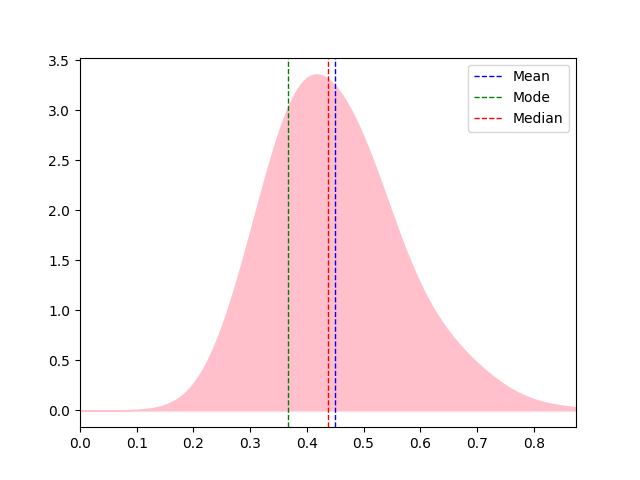}
    \caption{MSPT} \label{fig:pe_mspt}
  \end{subfigure}%
  \hspace*{\fill}   
  \begin{subfigure}{0.31\textwidth}
    \includegraphics[width=\linewidth]{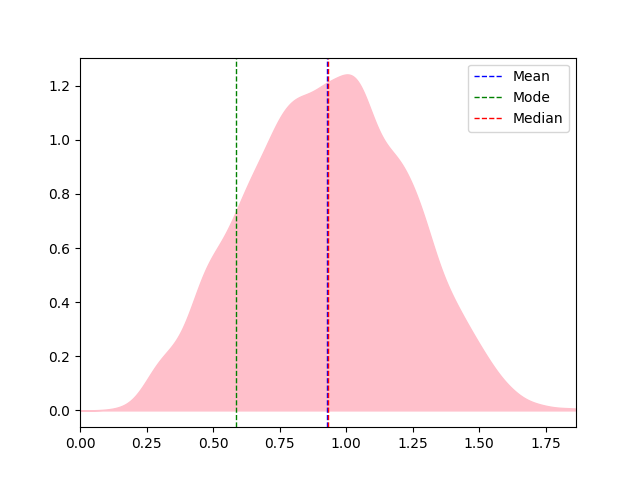}
    \caption{WLP} \label{fig:pe_wlp}
  \end{subfigure}%
  \hspace*{\fill}   
  \begin{subfigure}{0.31\textwidth}
    \includegraphics[width=\linewidth]{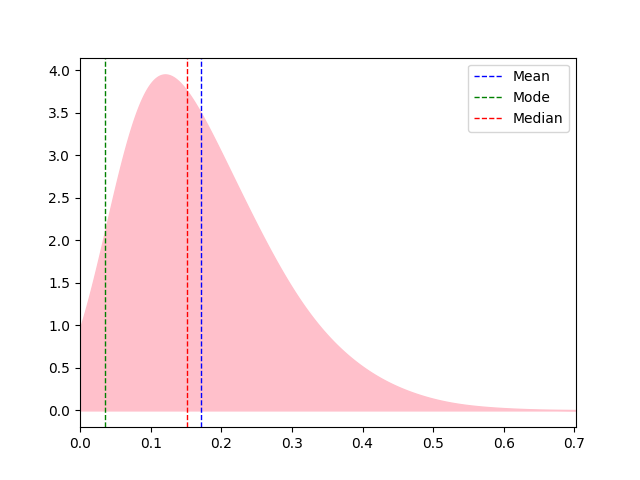}
    \caption{BC5CDR} \label{fig:pe_bc5}
  \end{subfigure}
\caption{The distribution of predictive entropy for each dataset.} \label{fig:pe_distribution}
\end{figure*}

Table \ref{tab:overall_result} shows the micro F1 score across datasets when $k = 8$ for those methods with in-context examples. 
First, it is notable that most GPT-based methods outperformed the fine-tuned BERT baseline. 
This indicates that GPT-3.5 may be a better choice when only a limited quantity of annotated data is available. 

Drilling down into the in-context example selecting methods results, it reveals that representativeness sampling cannot outperform random sampling. 
In contrast, KATE achieved improvements across datasets. 
This suggests that dynamic examples are more effective than static ones for in-context learning. 
Our proposed method, STAYKATE, which combined two types of examples, achieved the best performance.

Table \ref{tab:entitylevel_result} provides a more detailed, entity-level breakdown of the results shown in Table \ref{tab:overall_result}. 
Although the overall performance gap between KATE and STAYKATE in Table \ref{tab:overall_result} is relatively narrow, significant performance gaps can be observed among entity types within individual datasets on entity-level. 
In MSPT, \texttt{Operation} entities achieved a peak F1 score of 0.77, while \texttt{Property} entities only reached 0.25. 
Similarly, in WLP, \texttt{Action} entities attained a maximum F1 score of 0.82, contrasting with \texttt{Device} entities at 0.29.
These disparities may be attributed to the entity frequency within the corpus and the inherent complexity of entity types. 
Since \texttt{Operation} and \texttt{Action} entities are usually verbs, GPT-3.5 can easily extract them. 
In contrast, entities like \texttt{Materials}, \texttt{Property}, \texttt{Device} require more domain knowledge or context to accurately identify, increasing the difficulty. 
While most of the methods struggle with these domain-specific entities, the proposed STAYKATE method shows promising results. 
Furthermore, STAYKATE demonstrated more consistent performance across different entity types in comparison to some other methods, particularly in the WLP dataset. 
Additionally, the performance declined when combining random examples with KATE (Random + KATE) instead of combining examples from representativeness sampling, especially for the domain-specific entities.
This highlights the essential role of representativeness sampling in selecting static examples.
\begin{table}[t]
\scalebox{0.95}{
\begin{tabular}{l|cc|c}
\hline
\multirow{2}{*}{\textbf{Dataset}} & \multicolumn{1}{l}{\textbf{avg}} & \multicolumn{1}{l|}{\textbf{avg}}        & \multicolumn{1}{l}{\multirow{2}{*}{\textbf{Ratio}}} \\
                                  & \multicolumn{1}{r}{\# token}     & \multicolumn{1}{r|}{\# non-entity token} & \multicolumn{1}{l}{}                                \\ \hline
\textbf{MSPT}                     & 34.87                            & 26.66                                   & 76\%                                                \\ \hline
\textbf{WLP}                      & 15.87                            & 11.91                                    & 75\%                                                \\ \hline
\textbf{BC5CDR}                   & 25.39                            & 22.39                                    & 88\%                                                \\ \hline
\end{tabular}
}
\caption{The ratio of non-entity tokens.}
\label{tab:ratio_non-entity}
\end{table}

\begin{table*}[t]
\centering
\small
\begin{tabular}{cl|ccc|ccc|ccc}
\hline
\multicolumn{1}{l}{}             &                           & \multicolumn{3}{c|}{\textbf{Random}}                                & \multicolumn{3}{c|}{\textbf{KATE}}                                  & \multicolumn{3}{c}{\textbf{STAYKATE}}                              \\
\multicolumn{1}{l}{}             &                           & $k = 2$                & $k = 6$                & $k = 8$                 & $k = 2$                & $k = 6$                & $k = 8$                 & $k = 2$                & $k = 6$                & $k = 8$                \\ \hline
\textbf{Dataset}                 & \textbf{Entity (support)} & \multicolumn{1}{l}{} & \multicolumn{1}{l}{} & \multicolumn{1}{l|}{} & \multicolumn{1}{l}{} & \multicolumn{1}{l}{} & \multicolumn{1}{l|}{} & \multicolumn{1}{l}{} & \multicolumn{1}{l}{} & \multicolumn{1}{l}{} \\
\multirow{4}{*}{\textbf{MSPT}}   & Material (338)                 & 0.53                 & 0.55                 & 0.56                  & 0.58                 & 0.60                 & 0.60                  & 0.60                 & 0.60                 & \textbf{0.62}        \\
                                 & Operation (242)                 & 0.74                 & 0.76                 & 0.76                  & 0.75                 & 0.76                 & \textbf{0.77}         & 0.75                 & 0.76                 & 0.76                 \\
                                 & Property (105)                 & 0.07                 & 0.14                 & 0.15                  & 0.17                 & 0.22                 & 0.22                  & 0.19                 & 0.22                 & \textbf{0.25}        \\
                                 & micro avg                 & 0.52                 & 0.56                 & 0.57                  & 0.57                 & 0.60                 & 0.60                  & 0.58                 & 0.59                 & \textbf{0.61}        \\ \hline
\multirow{6}{*}{\textbf{WLP}}    & Action (275)                 & 0.77                 & 0.79                 & 0.79                  & 0.79                 & 0.81                 & \textbf{0.82}         & 0.79                 & 0.80                 & 0.81                 \\
                                 & Device (45)                  & 0.22                 & 0.24                 & 0.24                  & 0.24                 & 0.25                 & 0.25                  & 0.19                 & 0.25                 & \textbf{0.29}        \\
                                 & Location (122)                 & 0.46                 & 0.55                 & 0.54                  & 0.53                 & 0.55                 & \textbf{0.57}         & 0.56                 & 0.55                 & 0.55                 \\
                                 & Reagent (178)                 & 0.52                 & 0.62                 & 0.62                  & 0.63                 & 0.64                 & 0.66                  & 0.62                 & 0.66                 & \textbf{0.67}        \\
                                 & Seal (20)                  & 0.35                 & 0.48                 & \textbf{0.54}                  & 0.32                 & 0.39                 & 0.40                  & 0.41                 & 0.48        & 0.48        \\
                                 & micro avg                 & 0.59                 & 0.65                 & 0.65                  & 0.65                 & 0.67                 & 0.68                  & 0.65                 & 0.68                 & \textbf{0.69}        \\ \hline
\multirow{3}{*}{\textbf{BC5CDR}} & Chemical (291)                 & 0.70                 & 0.70                 & 0.71                  & 0.70                 & 0.71                 & 0.72                  & 0.72                 & 0.73                 & \textbf{0.74}        \\
                                 & Disease (185)                 & 0.63                 & 0.64                 & 0.65                  & 0.65                 & 0.65                 & 0.67                  & 0.66                 & 0.68                 & \textbf{0.70}        \\
                                 & micro avg                 & 0.67                 & 0.68                 & 0.69                  & 0.68                 & 0.69                 & 0.70                  & 0.70                 & 0.71                 & \textbf{0.72}        \\ \hline
\end{tabular}
\caption{Experimental results with various $k$ in-context examples for each dataset.} 
\label{tab:nshot_result}
\end{table*}

Another interesting point is that the performance of STAYKATE varied with different $\lambda$ values. 
We consider that the choice of $\lambda$ value depends on the characteristics of the dataset. 
Specifically, the ratio of non-entity tokens in the data pool. 
As shown in Table \ref{tab:ratio_non-entity}, the proportion of non-entity tokens for BC5CDR is higher than MSPT and WLP. 
Figure \ref{fig:pe_distribution} illustrates how this affects the distribution of predictive entropy for each dataset. 
For those datasets with lower non-entity token proportions, the distributions of predictive entropy were closer to normal distribution. 
However, for BC5CDR, the values were more concentrated in the low-value area. 
This is because the model typically has high confidence when predicting labels for non-entity tokens. 
Therefore, we suggest adjusting the selection zone using the parameter $\lambda$ when the proportion of non-entity tokens is high.

The experiment results with various $k$ in-context examples are given in Table \ref{tab:nshot_result}.
These results demonstrates that the extraction performance improved as $k$ increases. 
Comparing STAYKATE ($k = 8$, namely $k_s=2$ and $k_d=6$) with KATE ($k = 6$), STAYKATE outperformed KATE across most of the entities. 
This proves the effectiveness of the representativeness sampling that we combined in STAYKATE.

\section{Analysis}
\subsection{Error Analysis}
This section presents a comprehensive error analysis of GPT-3.5 outputs when $k=8$. 
The errors are broadly categorized into three types: overpredicting, oversight, and wrong entity type.

\begin{figure}[t]
  \centering
  \includegraphics[width=6.5cm]{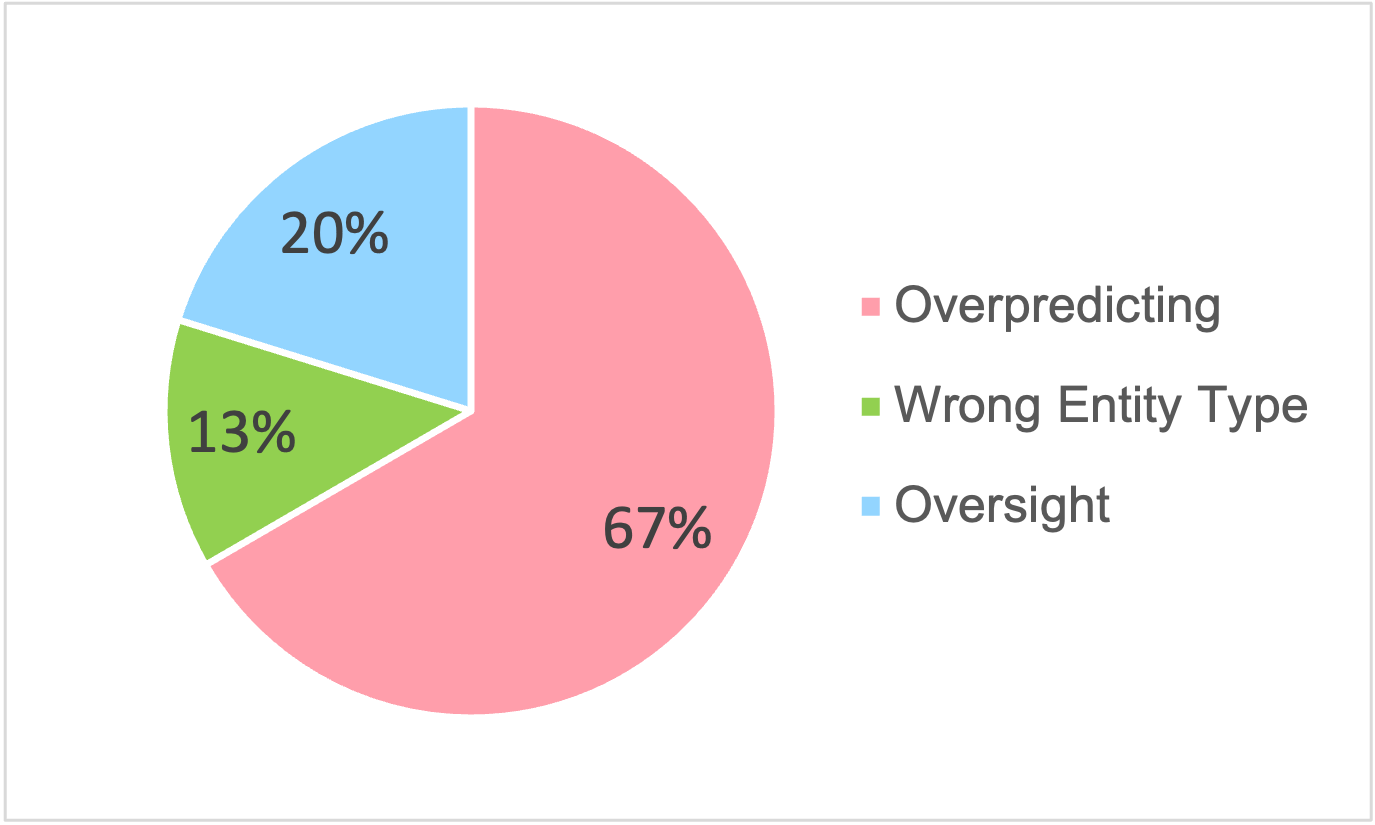}
   \caption{Statistics on the percentage of various error types.}
  \label{fig:error_analysis}
\end{figure}

\begin{figure}[t]
  \centering
  \includegraphics[width=6.5cm]{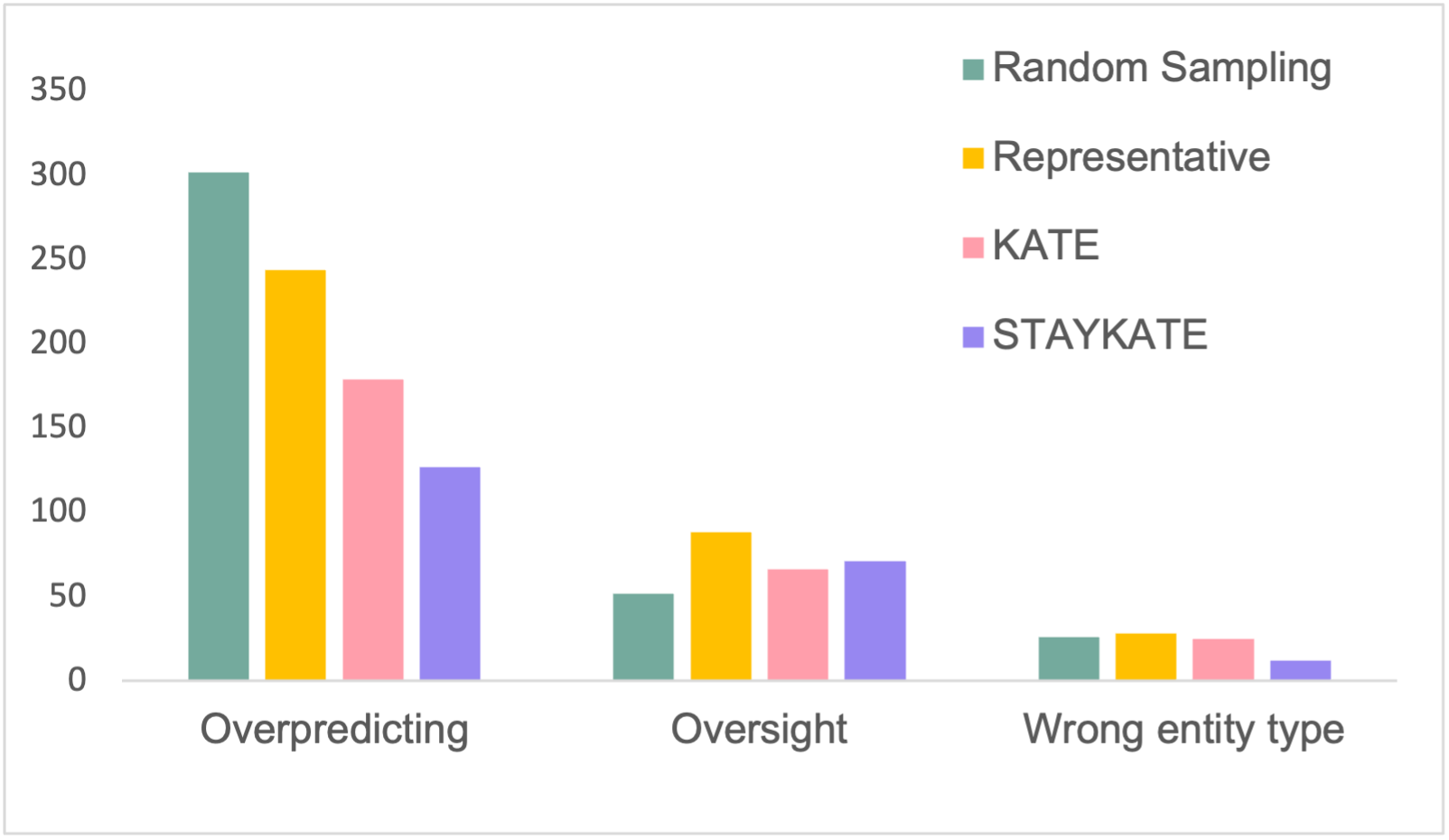}
  \caption{Statistics of errors across different selection methods for MSPT.}
  \label{fig:analysis_mspt}
\end{figure}

\begin{figure*}[t]
  \centering
  \includegraphics[scale=0.4]{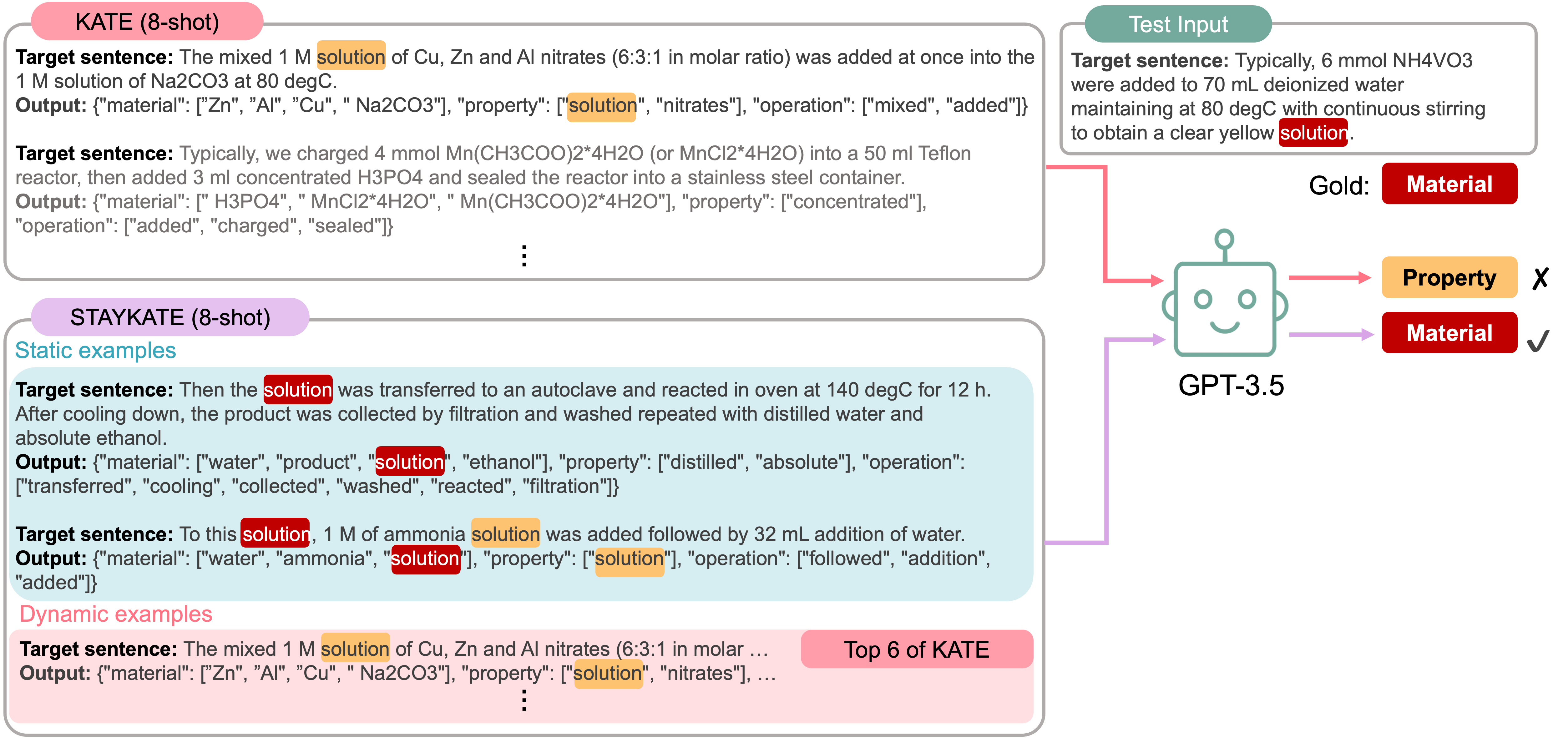}
  \caption{A case study comparing KATE with STAYKATE.}
  \label{fig:casestudy}
\end{figure*}

\paragraph{Overpredicting}
As shown in Figure \ref{fig:error_analysis}, overpredicting emerges as the most frequent issue in our experimental findings. 
This reveals that GPT-3.5 demonstrates a strong propensity to incorrectly assign pre-defined labels to non-entity tokens. 
This phenomenon also observed in other studies \cite{gptre, overpredict}. 
In our context, we found that the overpredicting occurred more frequently in \texttt{Property}, \texttt{Reagent}, and \texttt{Chemical}. 
The trend of these entity types to be overpredicted can be attributed to their context-dependent nature. 

\paragraph{Oversight}
In contrast to overpredicting, oversight occurred when GPT-3.5 treats a pre-defined label as a non-entity token. 
This error type was more frequently observed in entity types with low occurrence rates across the three datasets, such as \texttt{Device} and \texttt{Seal}. 
Due to the infrequent appearance of these entities, they are usually excluded from the selected examples, thus potentially misleading the models. 
Additionally, we observed that GPT-3.5 tends to ignore the abbreviations in \texttt{Material} and \texttt{Chemical} entities (e.g., original text: Fetal bovine serum (FBS); gold: Fetal bovine serum, FBS; predicted: Fetal bovine serum). 

\paragraph{Wrong entity type}
When GPT-3.5 misidentifies entity type A to entity type B, we call this wrong entity type. 
This error typically manifested in entities with the same surface representation but divergent contextual meanings. 
For example, in MSPT, \texttt{Property} and \texttt{Material} are usually misclassified. 
The words like ``powder'',``solution'', and ``suspension'' are easy to be confused since they can either be a physical substance or a descriptor of the characteristics of material.

\subsection{Why STAYKATE performs better?}
We checked the errors associated with different selection methods for MSPT. 
The analysis for WLP and BC5CDR can be found in Appendix \ref{app:more_analysis}. 
Figure \ref{fig:analysis_mspt} illustrates the statistics of three error types for Random Sampling, Representativeness Sampling, KATE, and STAYKATE. 
Notably, STAYKATE demonstrated a significant reduction in overpredicting errors. 
While there was a trade-off observed in the oversight errors, STAYKATE still exhibited comparable performance. 
Furthermore, regarding the wrong entity type errors, an issue that other selection methods struggled with, STAYKATE shows a lower rate. 

Additionally, we show one instance to highlight the benefits of hybrid in-context examples. 
As shown in Figure \ref{fig:casestudy}, the gold label for \textit{``solution''} in the test input is \texttt{Material}. 
The semantically similar examples retrieved by KATE offered some insights for identifying entities. 
However, in the context of the first example, \textit{``solution''} referred to \texttt{Property}, misleading GPT-3.5 to produce an incorrect response. 
In contrast, the two static representative examples provided by STAYKATE clarified the varying meanings of \textit{``solution''} across contexts, encouraging GPT-3.5 to better consider the nuance.

\section{Conclusions}
In this paper, we proposed a static-dynamic hybrid method, STAYKATE (STAtic\&dYnamic Knn-Augmented in-conText Example selection) to select the efficient in-context examples for scientific NER.
The experimental results indicated that under the low-resource setting, GPT-3.5 with ICL can surpass the fine-tuned base models.
Additionally, STAYKATE outperformed other existing selection methods. 
The enhancement was more significant in domain-specific entities.
The detailed analysis also showed the capability of STAYKATE to mitigate typical errors like overpredicting.

\section{Limitations}
There are several limitations to consider in this paper. 
We just evaluated STAYKATE on three datasets from the scientific domain.
Moreover, we just reported results with GPT-3.5. Therefore we need to evaluate STAYKATE with other LLMs.
Experiments with a larger number of examples are also necessary.

\bibliography{custom}

\appendix

\section{Data Statistics}\label{app:data_statistics}
Table \ref{tab:data_statistics} shows the statistical details of datasets. We also give the size of the subsets we used in our experiments. 
\begin{table}[ht]
\centering
\scalebox{0.8}{
\begin{tabular}{l|lll}
\hline
                & \textbf{Train (subset)} & \textbf{Dev (subset)} & \textbf{Test (subset)} \\ \hline
\textbf{MSPT}   & 1,758 (175)             & 105                   & 157                    \\
\textbf{WLP}    & 8,581 (176)             & 2,859 (199)           & 2,861 (194)            \\
\textbf{BC5CDR} & 4,288 (184)             & 4,299 (198)           & 4,600 (197)            \\ \hline
\end{tabular}
}
\caption{Statistical details of datasets. The size of subsets for training sets refer to the labeled data we used for KATE and BERT's fine-tuning.} 
\label{tab:data_statistics}
\end{table}

\section{Hyperparameters of BERT}\label{app:hyper}
The hyperparameters used are the max length of 350, the batch size of 32, the learning rate of 2e-5, and 20 epochs of training. 
We apply the EarlyStopping mechanism to prevent overfitting. 

\begin{figure*}[t]
  \centering
  \includegraphics[scale = 0.5]{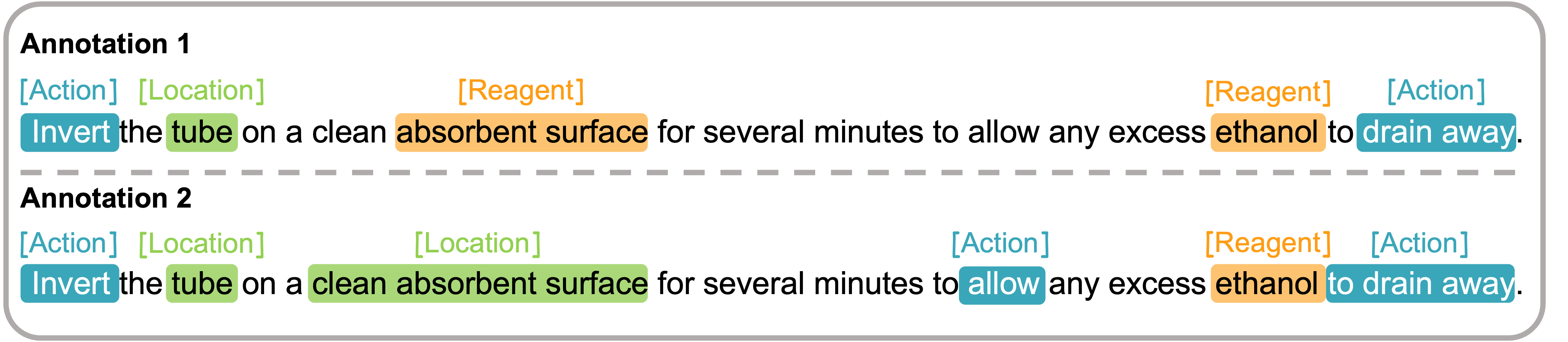}
  \caption{Sentence pair with inconsistent annotation in WLP.}
  \label{fig:annotation_wlp}
  
\end{figure*}
\begin{figure}[ht]
  \centering
  \includegraphics[width=\columnwidth]{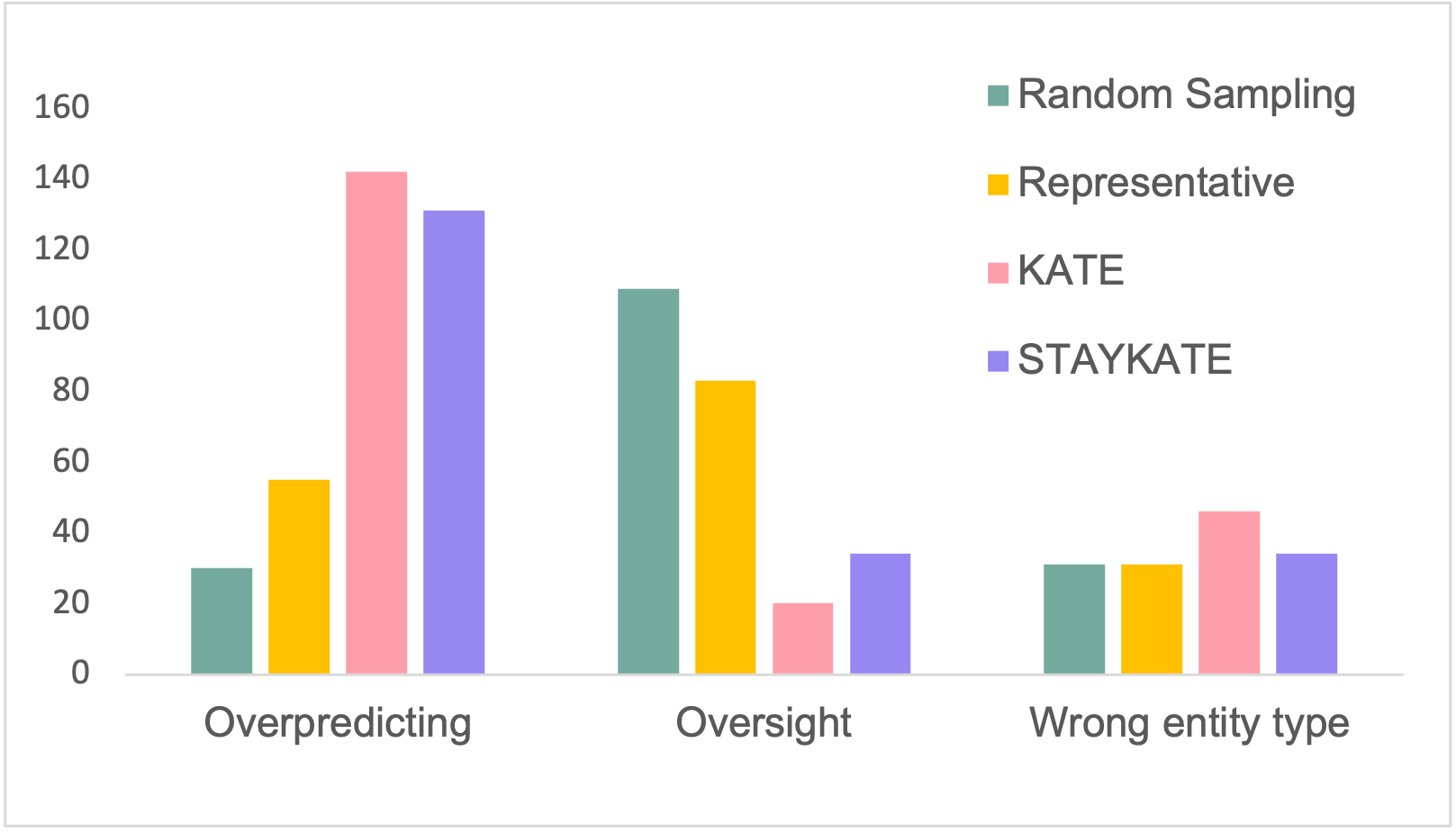}
  \caption{Statistics of errors across different selection methods for WLP.}
  \label{fig:analysis_wlp}
\end{figure}

\section{Statistics of errors for WLP and BC5CDR}\label{app:more_analysis}
Figure \ref{fig:analysis_wlp} shows the statistics of errors associated with different selection methods for WLP. 
Compared to the static-based methods (Random Sampling and Representativeness Sampling), both KATE and STAYKATE exhibit significantly higher overpredicting rate. 
We consider this can be attributed to the characteristics of the dataset. 
Since WLP consists of the instructions for biology and chemistry experiments, it contains more repetitive or extremely similar sentences. 
Moreover, we find that identical sentences in WLP can sometimes have varying annotations (See Figure \ref{fig:annotation_wlp}). 
The similarity-based retrieve model is likely to extract such sentence pairs. 
When a word in the same sentence is inconsistently labeled as an entity or non-entity, GPT-3.5 tends to assume it is an entity, leading to overpredicting.

\begin{figure}[ht]
  \centering
  \includegraphics[width=\columnwidth]{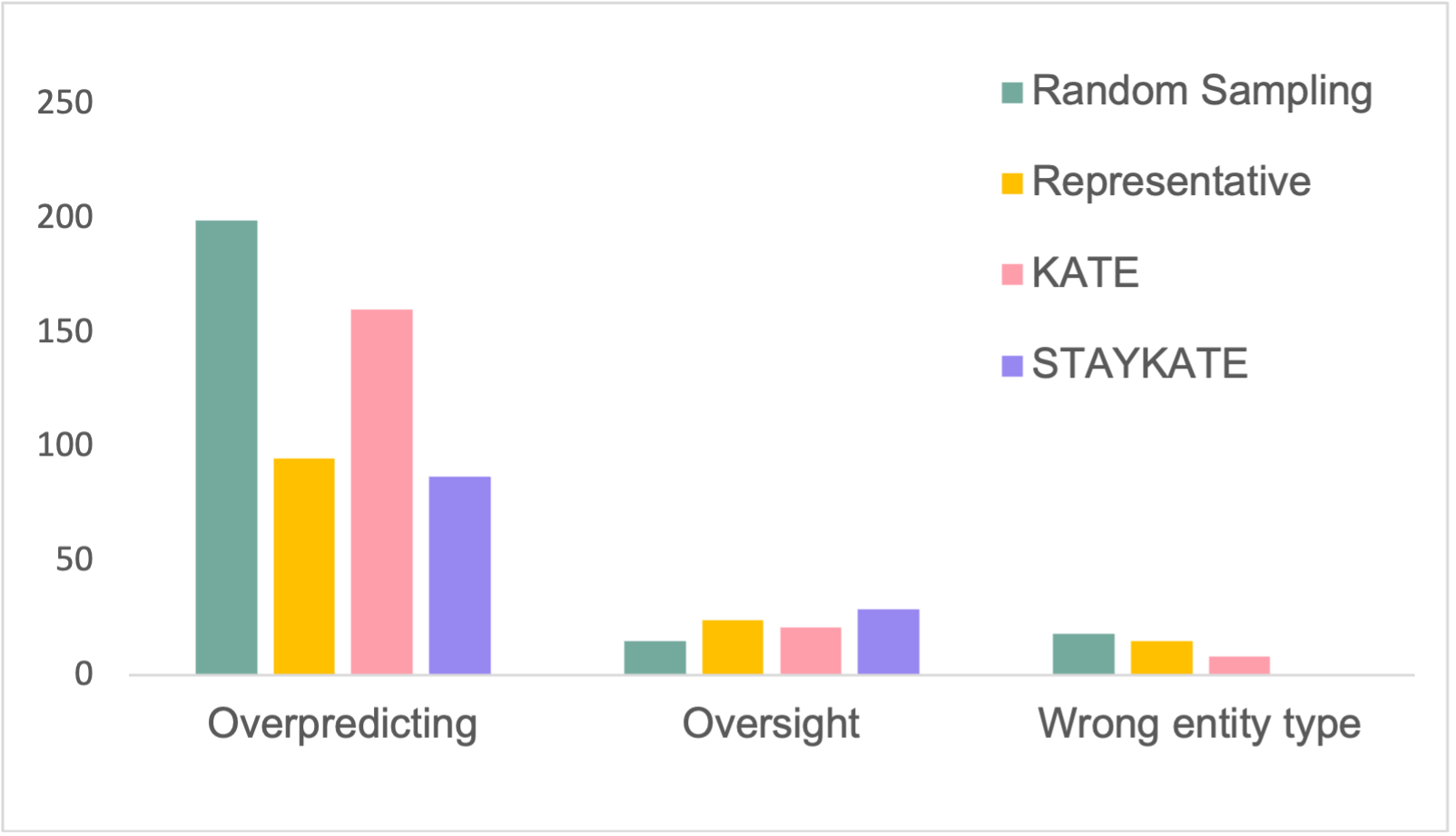}
  \caption{Statistics of errors across different selection methods for BC5CDR.}
  \label{fig:analysis_bc5}
\end{figure}

In the case of BC5CDR, similar trends are observed as in the MSPT dataset: STAYKATE demonstrates a reduction in both overpredicting errors and incorrect entity type classifications (Figure \ref{fig:analysis_bc5}). 
We consider that the mitigation in overpredicting owing to the STAYKATE provides examples that exclude entities. 
The wrong entity type errors often arise when a word is an abbreviation, making it challenging for the model to distinguish between chemical and disease entities. 
The in-context examples selected by STAYKATE also contain abbreviations, which offer crucial disambiguation cues.


\end{document}